\documentclass[conference]{IEEEtran}
\IEEEoverridecommandlockouts

\usepackage{fancyhdr}
\usepackage{cite}
\usepackage{amsmath,amssymb,amsfonts}
\usepackage{algorithmic}
\usepackage{graphicx}
\usepackage{textcomp}
\usepackage[table,xcdraw]{xcolor}
\usepackage{times}
\usepackage{epsfig}
\usepackage{lscape}
\usepackage{caption}
\usepackage{subcaption}
\usepackage[hidelinks]{hyperref}
\usepackage{censor}
\def\BibTeX{{\rm B\kern-.05em{\sc i\kern-.025em b}\kern-.08em
    T\kern-.1667em\lower.7ex\hbox{E}\kern-.125emX}}

\begin{document}

\makeatletter
\def\ps@IEEEtitlepagestyle{%
  \def\@oddfoot{\mycopyrightnotice}%
  \def\@oddhead{\hbox{}\@IEEEheaderstyle\leftmark\hfil\thepage}\relax
  \def\@evenhead{\@IEEEheaderstyle\thepage\hfil\leftmark\hbox{}}\relax
  \def\@evenfoot{}%
}
\def\mycopyrightnotice{%
  \begin{minipage}{\textwidth}
  \centering \tiny
  Copyright~\copyright~2022 IEEE. This paper has been accepted for publishing in Proceedings of 2022 Third International Conference on Intelligent Data Science Technologies and Applications (IDSTA, \url{https://intelligenttech.org/IDSTA2022/IDSTApackingList/26\_DTL2022\_RC\_8931.pdf}), and would be available via IEEE Xplore in the future. \\ Personal use of this material is permitted. Permission from IEEE must be obtained for all other uses, in any current or future media, including reprinting/republishing this material for advertising or promotional purposes, creating new collective works, for resale or redistribution to servers or lists, or reuse of any copyrighted component of this work in other works by sending a request to pubs-permissions@ieee.org.
  \end{minipage}
}
\makeatother

\title{EEG-based Image Feature Extraction for Visual Classification using Deep Learning}


\author{\IEEEauthorblockN{1\textsuperscript{st} Alankrit Mishra}
\IEEEauthorblockA{\textit{Department of Computer Science} \\
\textit{Lakehead University}\\
Thunder Bay, Canada \\
amishra1@lakeheadu.ca}
\and
\IEEEauthorblockN{2\textsuperscript{nd} Nikhil Raj}
\IEEEauthorblockA{\textit{Department of Computer Science} \\
\textit{Lakehead University}\\
Thunder Bay, Canada \\
nraj@lakeheadu.ca}
\and
\IEEEauthorblockN{4\textsuperscript{th} Garima Bajwa}
\IEEEauthorblockA{\textit{Department of Computer Science} \\
\textit{Lakehead University}\\
Thunder Bay, Canada \\
garima.bajwa@lakeheadu.ca}
}

\author{
\IEEEauthorblockN{
Alankrit Mishra\IEEEauthorrefmark{1},
Nikhil Raj\IEEEauthorrefmark{2}, and
Garima Bajwa\IEEEauthorrefmark{4}
}
\IEEEauthorblockA{
Department of Computer Science\\
Lakehead University\\
Thunder Bay, Ontario, Canada\\
Email: $\{$\IEEEauthorrefmark{1}amishra1,
\IEEEauthorrefmark{2}nraj,
\IEEEauthorrefmark{4}garima.bajwa$\}$@lakeheadu.ca
}
}

\maketitle

\begin{abstract}
While capable of segregating visual data, humans take time to examine a single piece, let alone thousands or millions of samples. The deep learning models efficiently process sizeable information with the help of modern-day computing. However, their questionable decision-making process has raised considerable concerns. Recent studies have identified a new approach to extract image features from EEG signals and combine them with standard image features. These approaches make deep learning models more interpretable and also enables faster converging of models with fewer samples. Inspired by recent studies, we developed an efficient way of encoding EEG signals as images to facilitate a more subtle understanding of brain signals with deep learning models. Using two variations in such encoding methods, we classified the encoded EEG signals corresponding to 39 image classes with a benchmark accuracy of 70\% on the layered dataset of six subjects, which is significantly higher than the existing work. Our image classification approach with combined EEG features achieved an accuracy of 82\% compared to the slightly better accuracy of a pure deep learning approach; nevertheless, it demonstrates the viability of the theory. 
\end{abstract}

\begin{IEEEkeywords}
EEG image encoding, computer vision, deep learning, ImageNet, multi-modal fusion, transfer learning, visual classification
\end{IEEEkeywords}

\section{Introduction}
Nowadays, digital data consists mainly of visual content such as images or videos. Visual classification is advancing our civilization through applications ranging from facial recognition to improved product discoverability. Humans have evolved to be natural and accurate classifiers, but our ability to categorize objects or create new categories is occasionally limited. We classify a scene using intuition and experience, but only if the distinctive patterns are visible \cite{5370737}. Machine perception can capture critical classifications and detect small patterns that the human mind ignores. Although deep learning models such as CNN deliver good performance, they lack clear explanations due to a black-box decision-making approach \cite{ul2019explaining}, time-consuming and computationally expensive for prediction and improving classification \cite{doan2013large}. 

Recently, Kaneshiro et al. identified a new approach to visual classification with machine learning using EEG signals from the human brain \cite{kaneshiro2015representational}. It attempts to map human perception to picture data collected from machine classifiers for visual classification tasks.

Previous studies have also used EEG signals as images, encoding them into space-time domain \cite{EEG_space_time} and time-frequency domain \cite{tsinalis2016automatic}. In this way, we can leverage the EEG cognitive features to aid in further classification of images by using techniques discussed in our methodology section. Representing these EEG signals in multidimensional encoded image space via a single sample provides rich data for classification. As deep learning models require a large amount of data to learn and extract features efficiently, these encodings enable us to do the same. 

In the following sections, we briefly describe the previous work, our initiatives, and the results with comparisons of various methodologies. The main focused approaches are:
\begin{enumerate}
    \item Visual classification using only images
    \item Visual classification using only EEG data
    \item Two-dimensional grayscale EEG encoded image data
    \item Visual classification using a multi-modal fusion of EEG and Image data
\end{enumerate}

\section{Contributions}
We found that EEG-ImageNet \cite{palazzo2020decoding} is one of the challenging benchmark datasets with 40 classes, which is a high number with EEG classification. Therefore, we can use this dataset to improve and achieve a robust classification of the EEG and image data. Our specific contributions using this data set are as follows:
\begin{itemize}
\item{
We modified an approach for encoding EEG data \cite{EEG_space_time} to an 8-bit grayscale image with 128 channels per trial. Using it with CNN + SVM pipeline-based transfer learning, we outperformed state-of-the-art models for EEG-ImageNet dataset classification.
}
\item {
We proposed two new methods, concatenation, and vertical stacking, using a multi-modal fusion of mixed input EEG and image features to explore and build brain-based visual features for future classification models.
}
\end{itemize}

\section{Related Work}
We reviewed the previous studies and approaches for EEG classification and visual classification using the EEG dataset.

EEG data consists of multiple channels of time-series signals per sample or trial. Over the years, many studies and state-of-the-art approaches have contributed to improve EEG data classifications. SyncNet \cite{SyncNet} and EEGNet \cite{lawhern2018eegnet}, used for benchmarking classes in the EEG datasets, are notable mentions of deep learning models of high performance.  

Li et al. \cite{SyncNet} built the SyncNet that used structured 1D convolution layers to extract power from both time and frequency domains and classified the data based on joint modeling of 1D CNNs. Lawhern et al. \cite{lawhern2018eegnet} used 2D CNNs along different dimensions of EEG data to create EEGNet. The first set learned frequency information via temporal convolution and then learned spatial features of specific frequency using a depth-wise set of CNNs.

One of the necessities in the standard EEG processing pipeline is feature engineering \cite{li2015eeg}. Traditional feature extraction provides only certain aspects of EEG, such as frequency or temporal domain content. A time-frequency resolution of EEG data can achieve the two-dimensional EEG representation. Therefore, the signals can be converted to a spectrogram image using STFT (short-time Fourier transform) \cite{tsinalis2016automatic, thodoroff2016learning} or to a scaleogram image using CWT (continuous wavelet transform) \cite{turk2019epilepsy}. Thus, it can leverage the performance of the pre-trained deep learning models using transfer learning. The earlier efforts by Raghu et al. \cite{RAGHU2020202} have shown success in EEG classification using spectrogram encoded images instead of raw EEG signals. Hence, we explored efficient ways to use the image-transformed features from EEG-ImageNet data in one of our classification experiments using CNN-based deep learning models without losing any channel or frequency information.

Zhang et al. \cite{EEG_space_time} followed a unique classification approach based on an EEG dataset \cite{begleiter_1999}. They used 8-bit heatmap scaling to convert the raw EEG signals into images. Later, they used pre-trained MobileNet to extract deep features from these images. In the end, they used an SVM classifier and obtained good classification performance.

Multi-modal fusion of diverse data has been emerging research to automate visual classification problems. Spampinato et al. \cite{spampinato2017deep} presented the first automated visual classification method driven by human brain signals using a CNN-based regression on the EEG manifold. Visual image stimuli evoked EEG data were learned with an RNN and then used to classify images into a learned EEG representation. Their promising results paved the way for human brain processes involved in effectively decoding visual recognition for further inclusion in automated methods.

Li et al. \cite{li2020perils} claimed that the results reported by Spampinato et al. \cite{spampinato2017deep} depended on a block design, and a rapid-event design process cannot replicate the results. The block design and training/test set splits were such that every trial in each test set came from a block with many attempts in the corresponding training set. Li et al. \cite{li2020perils} also claimed that the wrong block design approach led to a high classification accuracy from the long-term brain activity associated with a block rather than the perception of the class stimuli.

Palazzo et al.\cite{palazzo2020correct} defended their previous research \cite{spampinato2017deep} by counter analyzing the claims made by Li et al. \cite{li2020perils} while admitting the faults in data pre-processing. As a result, the classification performance was lower than the previously claimed average accuracy of around 83\% \cite{spampinato2017deep}. According to their latest work \cite{palazzo2020decoding}, the reduced accuracy was attributable to EEG drift because the earlier work mistakenly used unfiltered EEG data. The authors \cite{spampinato2017deep} achieved nearly 20\% accuracy with correctly filtered data (high-frequency gamma-band); EEGNet\cite{lawhern2018eegnet} reported about 30\% accuracy, and EEG-Channel Net\cite{palazzo2020decoding} obtained approximately 50\% accuracy. However, in their experimental finding, the temporal correlation in \cite{spampinato2017deep}'s data was nominal, and the block design was suitable for classification studies after pre-processing. Consequently, they corrected for the publicly available data with proper filtering.

\begin{table}[]
\caption{Performance comparison of previous approaches on EEG-ImageNet\cite{palazzo2020decoding} dataset.}
\label{tab:LR_table}
\resizebox{\columnwidth}{!}{%
\begin{tabular}{|ccc|}
\hline
\multicolumn{3}{|c|}{\cellcolor[HTML]{D9D9D9}Model performances with unfiltered EEG-ImageNet}                                                        \\ \hline
\multicolumn{1}{|c|}{\textbf{Classifier model}}  & \multicolumn{2}{c|}{\textbf{Accuracy on unfiltered  EEG-ImageNet data}}                           \\ \hline
\multicolumn{1}{|c|}{Stacked LSTMs\cite{spampinato2017deep}}              & \multicolumn{2}{c|}{0.83}                                                                         \\ \hline
\multicolumn{1}{|c|}{Cogni-Net\cite{mukherjee2019cogni}}                  & \multicolumn{2}{c|}{0.89}                                                                         \\ \hline
\multicolumn{1}{|c|}{Bi-LSTMs\cite{fares2018region}}                   & \multicolumn{2}{c|}{0.97}                                                                         \\ \hline
\multicolumn{1}{|c|}{LSTM-B\cite{zheng2020ensemble}}                     & \multicolumn{2}{c|}{0.97}                                                                         \\ \hline
\multicolumn{3}{|c|}{\cellcolor[HTML]{D9D9D9}Accurac on correctly filtered EEG-ImageNet}                                                              \\ \hline
\multicolumn{1}{|c|}{\textbf{Classifier models}} & \multicolumn{1}{c|}{\textbf{Accuracy on ({[}14-70{]} Hz)}} & \textbf{Accuracy on ({[}5-95{]} Hz)} \\ \hline
\multicolumn{1}{|c|}{Stacked LSTMs\cite{spampinato2017deep,palazzo2020decoding}}              & \multicolumn{1}{c|}{NA}                                    & 0.22                                 \\ \hline
\multicolumn{1}{|c|}{SyncNet\cite{SyncNet}}                    & \multicolumn{1}{c|}{0.24}                                  & 0.27                                 \\ \hline
\multicolumn{1}{|c|}{EEGNet\cite{lawhern2018eegnet}}                     & \multicolumn{1}{c|}{0.34}                                  & 0.32                                 \\ \hline
\multicolumn{1}{|c|}{EEG-ChannelNet\cite{palazzo2020decoding}}             & \multicolumn{1}{c|}{0.41}                                  & 0.36                                 \\ \hline
\multicolumn{1}{|c|}{GRUGate  Transformer\cite{GRUTarnsformer}}       & \multicolumn{1}{c|}{0.48}                                  & 0.46                                 \\ \hline
\end{tabular}%
}
\end{table}

Following this revelation, the performance of the models developed in \cite{fares2018region}, \cite{mukherjee2019cogni}, \cite{kavasidis2017brain2image} and \cite{zheng2020ensemble} cannot be compared as they are based on the  unfiltered EEG data from \cite{spampinato2017deep} and the filtered dataset was published in 2020 \cite{palazzo2020decoding}. 

Tao et al. \cite{GRUTarnsformer} used the filtered data from \cite{palazzo2020decoding} with different frequency sets of [55-95] Hz, [14-70] Hz, and [5-95] Hz to compare all state-of-the-art models. They based their proposed model for EEG classification on the GRUGate Transformer. They achieved 61\% accuracy on the high gamma band filtered data but reached only 49\% with all band frequency data ([5-95] Hz). Table \ref{tab:LR_table} compared the performance of all the previous studies that used the unfiltered as well as the correctly filtered EEG-ImageNet dataset.

\section{Dataset}
For this study, we have used the updated filtered dataset published in 2020 \cite{spampinato2017deep} \cite{palazzo2020decoding}. It is the first EEG dataset for ImageNet visual classification's multi-class subset. For future convenience, we will refer to this dataset as EEG-ImageNet.

EEG signals were recorded from six subjects viewing a subset of the ImageNet dataset with 40 classes, each containing 50 images. The EEG data was collected from 128 electrodes with a sampling rate of 1000 Hz and 500 ms in duration. According to Kaneshiro et al. \cite{kaneshiro2015representational}, the first 500 ms of single-trial EEG responses are informative for the categories and characteristics of visual objects in this investigation. They also found that as little as 80 milliseconds of response from a single electrode is enough to classify EEG signals.


\begin{figure*}[ht]
    \includegraphics[width=\textwidth]{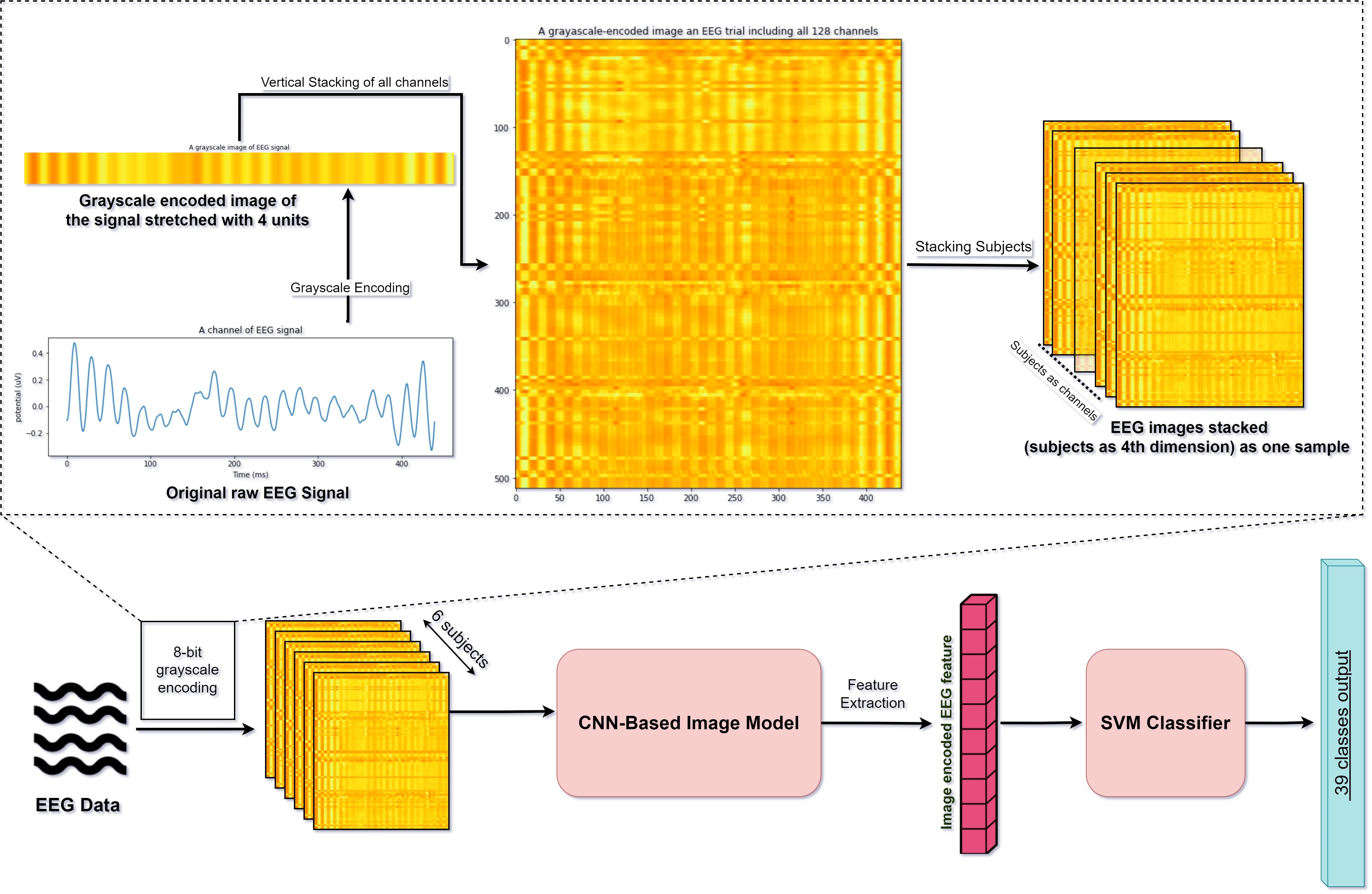}
    \centering
    \caption{The process of encoding EEG trials to images for EEG-to-Image-based models.}
    \label{fig:gem_model}
\end{figure*}

The number of trials found in the dataset is 11,964, after removing 36 low-quality samples from 12000 recordings. It is also worth mentioning that we discovered 11 missing trails for one class (mushrooms, labeled 33 in the dataset) for subject 1. We deleted all data with label 33 from the Image and EEG datasets, resulting in a 39-class dataset with 11,682 samples.

Many versions of the dataset were constructed with different bandpass filters, ranging from [5-95] Hz to [14-70] Hz for various experiments. For our research, we used both forms of filtered data to utilize various brain signal bands (theta, alpha, beta, and gamma) information captured during visual stimulation. Data was also adjusted using a z-score per channel to provide zero-centered values with a unitary standard deviation \cite{palazzo2020decoding}.

\section{Methodology} \label{Method}

\subsection{LSTM-based EEG Model} \label{EEG_raw_1}
The EEG data is a time series signal, so LSTM models are a reasonable choice to extract features for this application \cite{lawhern2018eegnet,fares2018region}. They can successfully learn on data with long-range temporal dependencies considering the time lag between inputs and their corresponding outputs. 

We used a mix of common stacked Bi-LSTMs and LSTMs to measure the baseline accuracy of our EEG data for image classification. Previous studies also showed comparable performance using stacked LSTMs \cite{spampinato2017deep} and stacked bi-directional LSTMs\cite{fares2018region} for EEG-ImageNet data.

\subsection{CNN-based Image Model}
Convolutional neural networks are the most effective deep learning models for extracting detailed features from images. With the support of ImageNet dataset \cite{deng2009imagenet} and deep learning models such as AlexNet, VGG \cite{VGG16}, ResidualNet \cite{ResNet}, MobileNet \cite{MobileNets}, and EfficientNet \cite{EfficientNet}, image classification has improved immensely and almost achieved its peak performance. However, the depth and parameters of these models necessitate a considerable resource for training with the ImageNet dataset. We can utilize these models on the go because they are already pre-trained using ImageNet weights. We took these pre-trained models and added a fully connected layer to fine-tune the model concerning our dataset, as our image data are a subset of ImageNet. These models provided us with the baseline classification performance of the 39 image classes in our study.

\subsection{EEG-to-Image-based model} \label{EEG_image_1}
Although deep learning models such as LSTMs and CNNs can extract features from raw time-series data directly, it is crucial to account for the noise and volatility in stochastic signals like the EEG. Additionally, before training the model with a sample, a pipeline technique should be determined to treat the raw EEG data as a feature consistent with the model's design. 

We designed a feature extractor method to transform the EEG signals into an 8-bit grayscale heatmap image \cite{EEG_space_time}. Having the flexibility to adjust the size of the signal image, we combined the EEG representations from all the 128 electrodes of a single trial. Thus, we created a unique signature for each sample without exhausting computational resources. This strategy allows integration of structural and textural analysis methods, such as pixel variance, morphological gradient calculations, normalization, and enhancement algorithms, to improve classification accuracy. It also permits the application of feature extraction methods that characterize the many forms, textures, and structures of each image, such as the Gray-Level Co-Occurrence Matrix (GLCM) \cite{haralick1973textural}, Hu's Moments\cite{hu1962visual}, and Local Binary Patterns \cite{ojala2002multiresolution}. Figure \ref{fig:gem_model} illustrates our approach to encoding an EEG trial to an image and subsequent design of a classification model.

\begin{figure}[h!]
    \includegraphics[width=\columnwidth]{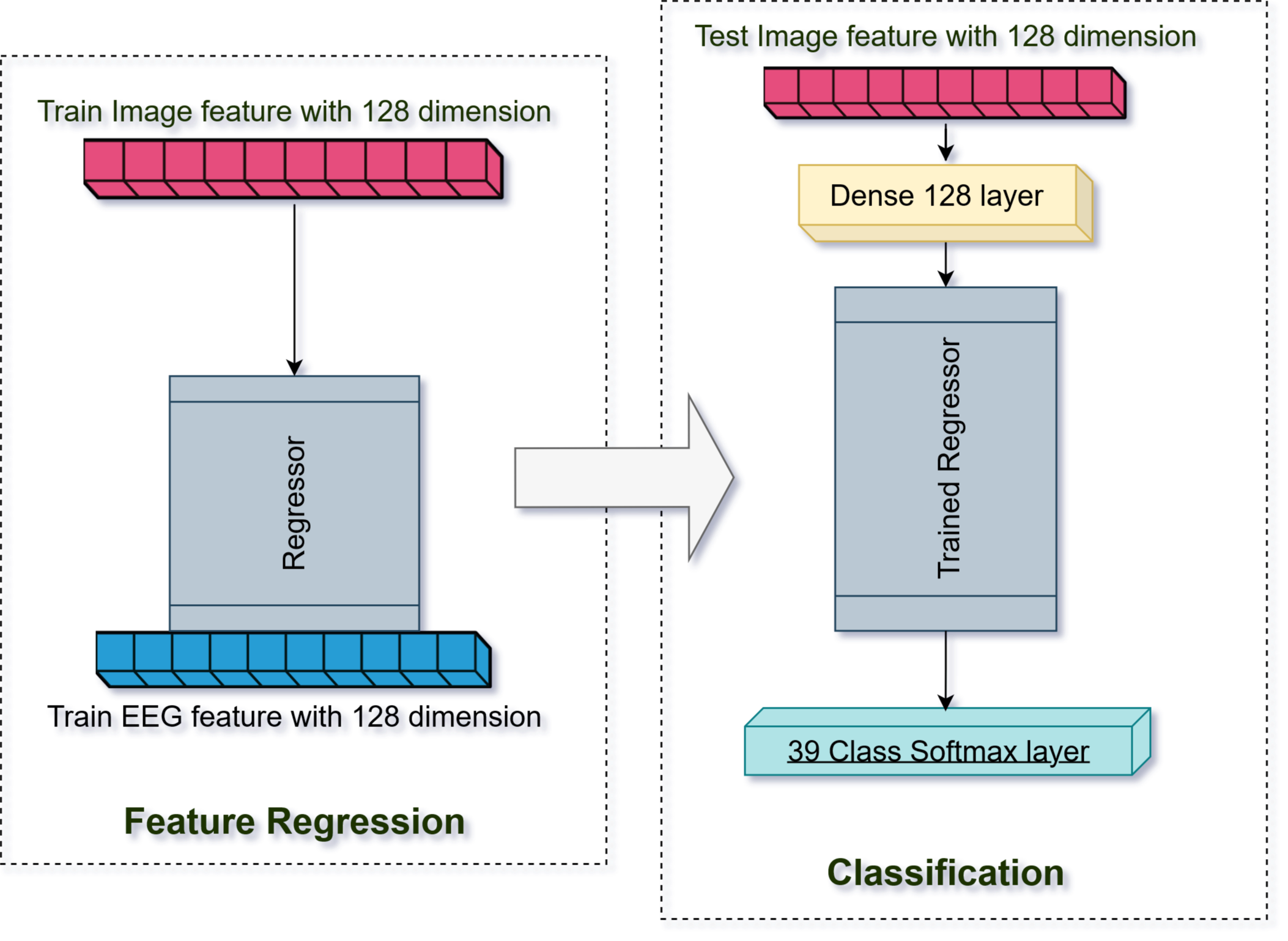}
    \centering
    \caption{The process used to build a Regression-based model.}
    \label{fig:regression}
\end{figure}

\subsection{Regression-based model}
The regression-based approach, similar to Spampinato et al. \cite{spampinato2017deep}, comprised a bi-directional LSTM to extract features from EEG signals and a CNN to extract image features for training. As a regressor, we employed a fully connected single-layer model to regress the image features derived from a pre-trained CNN model with EEG features acquired by the BiLSTM model. For simplicity, we set the feature dimension of both image and EEG features to 128. To match the number of EEG samples with the image samples (1947), we averaged the EEG signals between the six subjects for each image stimulus. For the test dataset, the regressor predicted or mapped the image features to the EEG features associated with the test images before using them in the classification model. In other words, EEG signal features facilitated learning of regressed image features, enabling image classification with these predicted characteristics. Figure \ref{fig:regression} illustrates the regression-based model approach.

\subsection{Concatenation-based model}

\begin{figure}[h!]
    \includegraphics[width=\columnwidth]{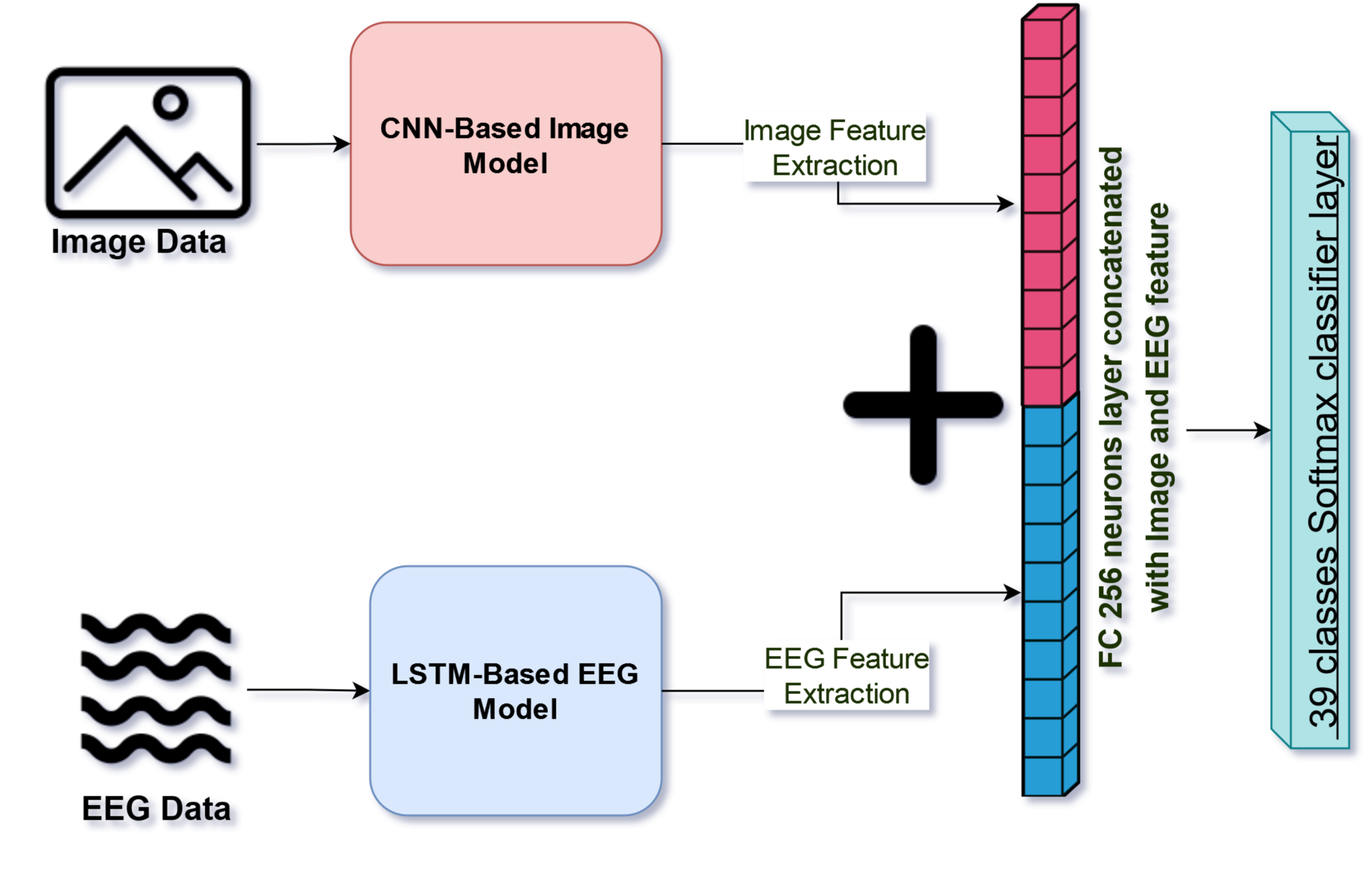}
    \centering
    \caption{The Concatenation-based model built using the baseline EEG and Image models.}
    \label{fig:feat_concat}
\end{figure}

A concatenation-based strategy typically combines the features collected from two or more machine learning models and then classifies those features into their corresponding labels. Our study has two classifier models: an LSTM-based EEG model and a CNN-based image model. We merge the penultimate levels of the models, that is, fully connected layers, right before the classification layer and then create a new classification layer to predict the 39 different classes in our dataset (a subset of ImageNet), shown in Figure \ref{fig:feat_concat}. Concatenated models often converge faster since they contain many parameters that help the final classification.

\subsection{Vertical Stacking model}
\begin{figure}[h!]
    \includegraphics[width=\columnwidth]{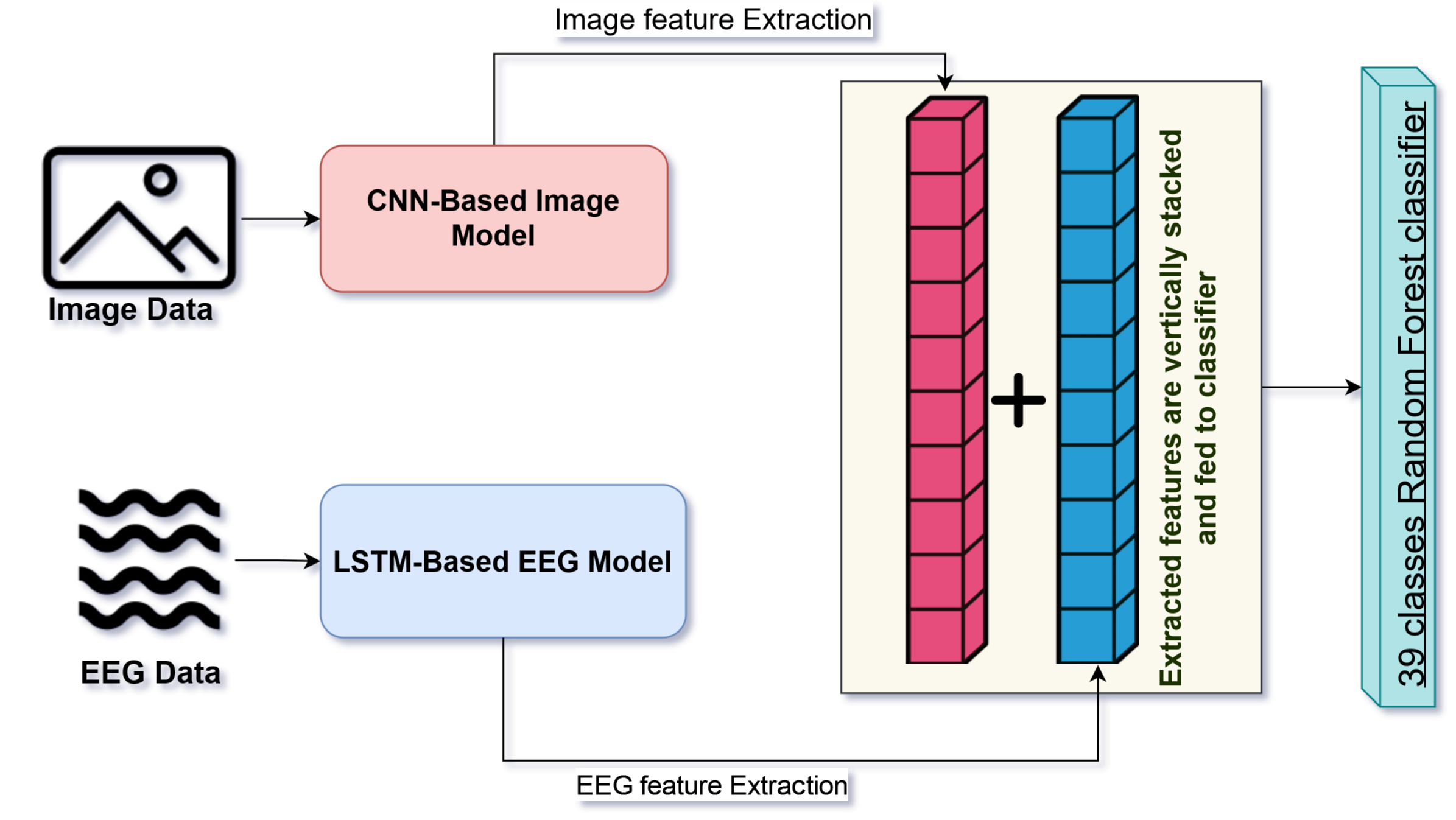}
    \centering
    \caption{The Vertical Stacking model obtained with stacked features from baseline models.}
    \label{fig:vert_stack}
\end{figure}

In this strategy, we constructed a new, expanded feature dataset by vertically stacking the features extracted from the EEG and image datasets and then assessed the overall classification accuracy using different classifier models (Figure \ref{fig:vert_stack}). The combined features contain the deep features retrieved from the baseline models (LSTM-based EEG and CNN-based Image models). We could append this data due to the dimensionality match in the EEG and image deep features (128). The goal was to augment the number of features from diverse data inputs with comparable properties.

\section{Experiments and Results} \label{experiment}
The length of an EEG sequence in the filtered EEG-ImageNet dataset was 500 ms. We selected 440 time points (20 - 460 ms) from 500 ms data, since the precise duration of each signal can vary \cite{palazzo2020decoding}. Thus, we excluded the beginning and final 20 samples (20 ms) to prevent interference in adjacent signal recordings. 

We used only 1947 out of 1996 image samples in the ImageNet subset after removing label 33 (mushrooms) associated with missing trials. Hence, the new size of the EEG-ImageNet dataset had 11,682 recordings (1947x6) with 39 class labels. We split the data into 70\% train, 15\% validation, and 15\% test sets. We stratified the data by image samples and labels, implying that each data split included trials from all subjects with the same visual stimulus in the same group. This type of stratification eliminated any bias in the split and avoided over-fitting during training.

Furthermore, we processed our EEG-ImageNet data based on the following experiments using the models described in Section \ref{Method}.

\subsection{Visual classification using only images}
The EEG-ImageNet dataset contains visual stimuli as a subset of the ImageNet dataset (1947 images from 39 classes). To measure the benchmark classification performance of this subset of ImageNet, we performed image classification experiments using different CNN-based pre-trained models such as AlexNet, VGG16, Resnet50, and MobileNet. We did not choose any model with high depth or parameters, as it would increase the complexity of the model. Unlike Palazzo et al. \cite{palazzo2020decoding}, we did not perform any image augmentation because we chose these models as image feature extractors for our future multi-modal implementations. 

\textit{Parameters:}
This model architecture consisted of an input layer of shape (224x224x3), followed by a functional model layer that fits CNN-based pre-trained models, a dense layer of 128 neurons and a softmax class layer as classifiers. We used a stochastic gradient descent optimizer to train the data.

\textbf{Results:}
We found that the ResNet50 model performed better than other CNN-based models with a test accuracy of 84\% as it converges faster for the small number of samples per class \cite{zheng2020ensemble}.

\begin{table*}[]
\centering
\caption{Performance comparison of classification models with varying cut-off frequencies in bandpass filters on the EEG-ImageNet data.}
\label{tab:eggdata_comparison}
\resizebox{\textwidth}{!}{%
\begin{tabular}{|l|l|c|c|}
\hline
\rowcolor[HTML]{C0C0C0} 
\textbf{EEG data Encoding}                                                                                            & \multicolumn{1}{l|}{\cellcolor[HTML]{C0C0C0}\textbf{Classifier models}} & \textbf{\begin{tabular}[c]{@{}c@{}}Accuracy on beta-gamma\\  filtered data ({[}14-70{]} Hz)\end{tabular}} & \textbf{\begin{tabular}[c]{@{}c@{}}Accuracy on All freq.\\ data ({[}5-95{]} Hz)\end{tabular}} \\ \hline
\rowcolor[HTML]{EFEFEF} 
Raw EEG data                                                                                                          & Stacked LSTMs\cite{spampinato2017deep}\cite{palazzo2020decoding}                                                          & NA                                                                                            & 0.22                                                                                     \\ \hline
\rowcolor[HTML]{EFEFEF} 
Raw EEG data                                                                                                          & SyncNet\cite{SyncNet}                                                                 & 0.24                                                                                          & 0.27                                                                                     \\ \hline
\rowcolor[HTML]{EFEFEF} 
Raw EEG data                                                                                                          & EEGNet\cite{lawhern2018eegnet}                                                                  & 0.34                                                                                          & 0.32                                                                                     \\ \hline
\rowcolor[HTML]{EFEFEF} 
Raw EEG data                                                                                                          & EEG-ChannelNet\cite{palazzo2020decoding}                                                          & 0.41                                                                                          & 0.36                                                                                     \\ \hline
\rowcolor[HTML]{EFEFEF} 
Raw EEG data                                                                                                          & GRUGate Transformer\cite{GRUTarnsformer}                                                     & 0.48                                                                                          & 0.46                                                                                     \\ \hline
\rowcolor[HTML]{ECF4FF} 
Raw EEG data                                                                                                          & LSTM based Model (\ref{EEG_raw_1},\ref{EEG_image_raw})                                                        & 0.28                                                                                          & 0.26                                                                                     \\ \hline
\rowcolor[HTML]{ECF4FF} 
\textbf{Grayscale image encoded EEG data} (\ref{EEG_image_1},\ref{EEG_image_raw})                                                                           & \textbf{EfficientNet +  SVM(rbf)}                                      & \textbf{0.51}                                                                                 & \textbf{0.64}                                                                            \\ \hline
\rowcolor[HTML]{ECF4FF} 
\textbf{\begin{tabular}[c]{@{}l@{}}Grayscale image encoded EEG data \\ with subjects as channels (six)
\end{tabular}} (\ref{EEG_image_1},\ref{EEG_image_raw}) & \textbf{EfficientNet +   SVM(rbf)}                                      & \textbf{0.68}                                                                                 & \textbf{0.70}                                                                            \\ \hline
\end{tabular}%
}
\end{table*}

\subsection{Visual classification using only EEG data} \label{EEG_image_raw}
In this section, we describe our classification experiments performed with EEG-ImageNet data.
\vspace{0.1cm}
\subsubsection{Raw EEG data with LSTM-based EEG Model}
We directly used raw time series EEG signals from all subjects. The shape of each input EEG data sample is (440x128), where 440 is the number of time points and 128 is the number of channels for each trial.  

\textit{Parameters:}
The LSTM-based EEG Model was built with an input layer with the same shape as each EEG sample. It was connected to 50 stacked bidirectional LSTMs, followed by two stacks (128 and 50) of common LSTMs, and finally, a dense layer of 128 neurons. We used the adam optimizer to train the model with a softmax classifier.

\textbf{Results:} We observed the performance our LSTM-based EEG Model followed a similar trend (Table \ref{tab:eggdata_comparison}) as previous state-of-the-art models \cite{SyncNet,lawhern2018eegnet,palazzo2020decoding,GRUTarnsformer} using only the raw EEG signals. The beta-gamma filtered data [14-70] Hz showed somewhat better performance than all frequency band data [5-95] Hz.

\vspace{0.2cm}
\subsubsection{EEG data encoded as 2D vectors with EEG-to-Image-based model}
We performed two kinds of processing in the EEG-ImageNet dataset for EEG signal-to-image models. 

In the first method, we created a grayscale heatmap of the EEG signals for every subject for each trial (40 trials/images per class). The process involved normalizing the signals with a min-max scalar to transform values in the range of (0,1). The normalized signals were then converted to 8-bit grayscale heatmap images using an encoding scheme described by Zhang et al. \cite{EEG_space_time}. However, we used a factor of four instead of 32 to increase pixel values to incorporate data from all 128 electrodes of a single trial. After this, we vertically layered each electrode's (4, 440) grayscale image to create an image of size (512, 440) corresponding to all 128 channels. In the end, each EEG trial's grayscale image was cloned three times (512x440x3) and resized to (224x224x3) to match the input shape of the pre-trained models such as MobileNet, Resnet, and EfficientNet.

We opted to group all subjects' EEG signals corresponding to the same image stimulus in the second method. As a result, the dimension of our EEG encoded image changed from (512x440x3) having 11,682 trials to (512x440x6) with 1947 trials. Instead of replicating the grayscale image data three times, we used the trails of six subjects for the same image as channels. The inference is that this will improve the efficiency of data input processing.

\textit{Parameters:} 
We leveraged a pipeline framework for transfer learning to train our models with EEG-encoded image data. We extracted the deep features from 8-bit grayscale images of each EEG trial using CNN-based image models such as MobileNet, ResNet, and EfficientNet. These deep features were then input into various machine learning classifiers, including SVM (RBF kernel), K Nearest Neighbor, Random Forest, Decision Tree, and Logistic Regression, to assess our classification performance.

\begin{table}[h]
\caption{Classification accuracy of different CNN (3 channels) + ML classifier models on grayscale EEG encoded image data for [14-70] Hz data.} 
\label{tab:space_time_14_70}
\resizebox{\columnwidth}{!}{%
\begin{tabular}{|l|c|c|}
\hline
\rowcolor[HTML]{E7E6E6} 
\textbf{CNN Extractor + classifier} & \textbf{\begin{tabular}[c]{@{}c@{}}Image size \\ (512x440)\end{tabular}} & \textbf{\begin{tabular}[c]{@{}c@{}}Image resized \\ (224x224)\end{tabular}} \\ \hline
MobileNet +   SVM(rbf)              & 0.42                                                                        & 0.36                                                                           \\ \hline
MobileNet +   kNN                   & 0.41                                                                        & 0.36                                                                           \\ \hline
ResNet +   SVM(rbf)                 & 0.5                                                                         & 0.43                                                                           \\ \hline
ResNet + kNN                        & 0.49                                                                        & 0.41                                                                           \\ \hline
\textbf{EfficientNet +   SVM(rbf)}  & \textbf{0.51}                                                               & \textbf{0.41}                                                                  \\ \hline
EfficientNet +   kNN                & 0.5                                                                         & 0.41                                                                           \\ \hline
\end{tabular}%
}
\end{table}

\begin{table*}[!htb]
\caption{The results of multi-modal fusion experiments compared with performance of the baseline models.}
\centering
\resizebox{\textwidth}{!}{%
\begin{tabular}{|c|l|l|c|c|}
\hline
\rowcolor[HTML]{C0C0C0} 
\textbf{Exp.} & \textbf{Implementation approach}                                   & \textbf{Model Used}                              & \textbf{Dataset \cite{palazzo2020decoding}} & \textbf{Accuracy} \\ \hline
1             & LSTM-based EEG Model (LEM)                                              & Stacked (BiLSTM + LSTMs) and 128 FC                                & EEG                                                                 & 0.28              \\ \hline
2             & CNN-based Image Model (CIM)                                            & Resnet50 pretrained with FC 128                   & Image                                                               & 0.84              \\ \hline
3             & Vertical Stacking                                                  & Resnet50 pretrained and LEM (end to end)         & Image + EEG                                                         & 0.7               \\ \hline
4             & Regression-based Model \cite{spampinato2017deep} & LEM feature regressed with CIM     & Image + EEG                                                         & 0.03              \\ \hline
\textbf{5}    & \textbf{Concatenation-based Model}                                       & \textbf{LEM concatenated with CIM} & \textbf{Image + EEG}                                                & \textbf{0.82}     \\ \hline
\end{tabular}%
}
\label{tab:results}
\end{table*}

\textbf{Results:} 
We tested the accuracy of all top-pipeline combinations using [14-70] Hz EEG-ImageNet data encoded as grayscale images, shown in Table \ref{tab:space_time_14_70}. We noticed information loss from the encoded images when the image size reduced from (512x440) to (224x224). We also found that the EfficientNet feature extractor with SVM classifier outperformed other model combinations. Hence, we chose to run this model setup for all frequency data available (i.e., [5-95] Hz EEG-ImageNet data).

Ultimately, we compared the performance of state-of-the-art EEG classifiers with the classifier we designed in Table \ref{tab:eggdata_comparison}. The comparison shows the classification accuracy for both data types available for the EEG-ImageNet dataset. \textit{Our grayscale EEG encoded image approach trained with EfficientNet + SVM (RBF kernel) classifier achieved approximately 21\% higher accuracy than other approaches using the all frequency dataset [5-95] Hz.} It is worth mentioning that our other approaches also performed well with the filtered data.

\subsection{Visual classification using a multi-modal fusion of EEG and Image data}
This section explains the experiment involving a fusion of the spatial features extracted from Images with the cognitive features of visual stimuli from EEG-ImageNet data.

The Regression-based and Concatenation-based models used averaged EEG readings from all subjects for each image as the visual stimuli. The EEG data in EEG-ImageNet was six times the size of the Image data due to the usage of each image to record EEG responses from six participants. The sample disparity between the EEG-ImageNet (1947x6) and ImageNet (1947) datasets posed challenges to concurrently training both the LSTM-based EEG model and CNN-based Image model. Thus, to make the inputs of both models the same size, we averaged subjects' readings and normalized the feature-set using a standard scalar to aid convergence for the concatenation, and feature regression approaches. However, we did not require such an averaging in the Vertical Stacking approach model for deep features.

\textit{Parameters:} The parameters of the CNN-based Image model and LSTM-based EEG model remained the same for current fusion experiments as used previously. We used a softmax classifier for classification in the end for both, Regression-based and Concatenation-based models as the dataset has multiple classes (39). 
In the Vertical Stacking approach model, the stacked features from the LSTM-based EEG model and CNN-based Image Models were the input to various classifiers such as KNN, SVM, Random Forest, and Logistic Regression.

\textbf{Results:}
Table \ref{tab:results} displays the key findings of our multi-modal fusion techniques compared with baseline models (LSTM-based EEG and CNN-based Image models). As noted previously, the accuracy of the CNN-based Image model, when trained in an isolated condition, was 84 percent [ResNet50(pre-trained) + 128 Dense]. In contrast, the baseline accuracy for LSTM-based-EEG model [50 BiLSTM + (128 + 50)LSTMs + 128 Dense] was a mere 26\%. The vertical feature stacking method achieved a modest performance with an accuracy of 70\%. Furthermore, the performance of the Concatenation-based model was 82\%. The Regression-based technique's performance was significantly inferior to the baseline performance of the LSTM-based EEG model.

\section{Discussion}
LSTM \cite{lawhern2018eegnet} and 1D CNN-based models \cite{palazzo2020decoding} generally work well with time-series data including many EEG datasets \cite{begleiter_1999, DBLP:journals/corr/abs-1904-09111}. However, this typical strategy of using LSTM with EEGs did not yield a high classification accuracy with the EEG-ImageNet dataset. The low performance of the EEG-ImageNet dataset can be attributed to its complexity, as it is one of the most extensive EEG datasets available in terms of containing an unusually high number of classes (39) \cite{ahmed2021object}, thus a harder EEG classification problem.

2D CNNs can extract deep features from data very effectively compared to LSTMs and 1D CNNs. Similarly, EEG signals can be encoded into 2D time-frequency image representations (spectrograms/scaleograms) with methods like STFT and CWT to leverage the deep feature extraction capability of pre-trained CNNs (\cite{ResNet}\cite{EfficientNet}). However, more resources are required for computation as we need a separate time-frequency map of each of the 128 channels, increasing sample sizes and adding enormous complexity to the data processing. In addition to the complexity, STFT and CWT methods lose feature information when resizing the images produced from the encoding. Hence, the previous studies \cite{thodoroff2016learning}\cite{tsinalis2016automatic} chose a selected number of channels for each trial instead of all EEG electrodes. 

On the other hand, our method of encoding the EEG signals to grayscaled image vectors outperforms the current state-of-the-art methods significantly (21\%), as seen in Table \ref{tab:eggdata_comparison}. The reason is that we accommodate the two-dimensional feature information from all the 128 channels in a single image by stretching the feature space of each channel instead of compressing it. 

In a different approach, we consider the six subjects as six separate channels of an image because the CNN can accommodate more than three channels. This strategy reduces the redundancy of the six different readings from the subjects but preserves the essential visual stimulus information (all the subjects are watching the same image).

We also observe that the EEG data consisting of all frequencies, i.e., from 5 Hz to 95 Hz, performs better than the dataset filtered with beta and gamma bands, i.e., 14 Hz to 70 Hz when encoded with two-dimensional image representations. This higher performance shows that EEG data has definitive classifying information in alpha frequency and can be helpful when we encode the dataset into deeper dimensions.

As per our findings, while resilient for multi-modal fusion, the Regression-based model did not perform with the filtered dataset. Palazzo et al. \cite{palazzo2020correct} also evaluated the performance of the regression technique on the same dataset and obtained a similar comparable low accuracy. The high accuracy achieved in earlier studies with regression \cite{spampinato2017deep}\cite{zheng2020ensemble} was led by some extent of EEG drift, which caused a bias in signals \cite{palazzo2020decoding}.

However, when using features from the mixed data (EEG signals + image data) as inputs, the Concatenation-based Model materialized with reasonable accuracy. Although this model's accuracy is marginally reduced compared to the baseline CNN-based Image model, it is noteworthy as our findings aim to develop multi-modal fusion techniques to utilize brain-based visual features in future classification models.

%

\section{Conclusion}
Based on the performance of all the visual classification techniques described in our study, we conclude that encoding EEG data strategically into a two-dimensional feature space provides more exploratory information than raw EEG signals. Furthermore, we learn that the EEG data, presented as an input image to the deep learning models, the data from low-frequency EEG bands such as alpha are more accessible and contribute significantly to visual classification.

We also infer that the image classification can be enhanced and explained using multi-modal data with both EEG and images. The Concatenation-based model strategy for fusing the mixed input data performs better than all other fusion techniques. Thus, our attempt to map image data with EEG data using all of the subjects' trials improves the convergence and efficiency of the model.

In the future, we will explore additional techniques to efficiently encode EEG data for classification and enhance the multi-modal fusion approaches of EEG and Image data.

\bibliographystyle{IEEEtran}
\bibliography{references}

\begin{thebibliography}{10}
\providecommand{\url}[1]{#1}
\csname url@samestyle\endcsname
\providecommand{\newblock}{\relax}
\providecommand{\bibinfo}[2]{#2}
\providecommand{\BIBentrySTDinterwordspacing}{\spaceskip=0pt\relax}
\providecommand{\BIBentryALTinterwordstretchfactor}{4}
\providecommand{\BIBentryALTinterwordspacing}{\spaceskip=\fontdimen2\font plus
\BIBentryALTinterwordstretchfactor\fontdimen3\font minus
  \fontdimen4\font\relax}
\providecommand{\BIBforeignlanguage}[2]{{%
\expandafter\ifx\csname l@#1\endcsname\relax
\typeout{** WARNING: IEEEtran.bst: No hyphenation pattern has been}%
\typeout{** loaded for the language `#1'. Using the pattern for}%
\typeout{** the default language instead.}%
\else
\language=\csname l@#1\endcsname
\fi
#2}}
\providecommand{\BIBdecl}{\relax}
\BIBdecl

\bibitem{5370737}
J.~MacInnes, S.~Santosa, and W.~Wright, ``Visual classification: Expert
  knowledge guides machine learning,'' \emph{IEEE Computer Graphics and
  Applications}, vol.~30, no.~1, pp. 8--14, 2010.

\bibitem{ul2019explaining}
M.~ul~Hassan, P.~Mulhem, D.~Pellerin, and G.~Qu{\'e}not, ``Explaining visual
  classification using attributes,'' in \emph{2019 International Conference on
  Content-Based Multimedia Indexing (CBMI)}.\hskip 1em plus 0.5em minus
  0.4em\relax IEEE, 2019, pp. 1--6.

\bibitem{doan2013large}
T.-N. Doan, T.-N. Do, and F.~Poulet, ``Large scale visual classification with
  many classes,'' in \emph{International Workshop on Machine Learning and Data
  Mining in Pattern Recognition}.\hskip 1em plus 0.5em minus 0.4em\relax
  Springer, 2013, pp. 629--643.

\bibitem{kaneshiro2015representational}
B.~Kaneshiro, M.~Perreau~Guimaraes, H.-S. Kim, A.~M. Norcia, and P.~Suppes, ``A
  representational similarity analysis of the dynamics of object processing
  using single-trial eeg classification,'' \emph{Plos one}, vol.~10, no.~8, p.
  e0135697, 2015.

\bibitem{EEG_space_time}
\BIBentryALTinterwordspacing
H.~Zhang, F.~H.~S. Silva, E.~F. Ohata, A.~G. Medeiros, and P.~P.
  Rebouças~Filho, ``Bi-dimensional approach based on transfer learning for
  alcoholism pre-disposition classification via eeg signals,'' \emph{Frontiers
  in Human Neuroscience}, vol.~14, 2020. [Online]. Available:
  \url{https://www.frontiersin.org/article/10.3389/fnhum.2020.00365}
\BIBentrySTDinterwordspacing

\bibitem{tsinalis2016automatic}
O.~Tsinalis, P.~M. Matthews, Y.~Guo, and S.~Zafeiriou, ``Automatic sleep stage
  scoring with single-channel eeg using convolutional neural networks,''
  \emph{arXiv preprint arXiv:1610.01683}, 2016.

\bibitem{palazzo2020decoding}
S.~Palazzo, C.~Spampinato, I.~Kavasidis, D.~Giordano, J.~Schmidt, and M.~Shah,
  ``Decoding brain representations by multimodal learning of neural activity
  and visual features,'' \emph{IEEE Transactions on Pattern Analysis and
  Machine Intelligence}, vol.~43, no.~11, pp. 3833--3849, 2020.

\bibitem{SyncNet}
Y.~Li, K.~Dzirasa, L.~Carin, D.~E. Carlson \emph{et~al.}, ``Targeting eeg/lfp
  synchrony with neural nets,'' \emph{Advances in Neural Information Processing
  Systems}, vol.~30, 2017.

\bibitem{lawhern2018eegnet}
V.~J. Lawhern, A.~J. Solon, N.~R. Waytowich, S.~M. Gordon, C.~P. Hung, and
  B.~J. Lance, ``Eegnet: a compact convolutional neural network for eeg-based
  brain--computer interfaces,'' \emph{Journal of neural engineering}, vol.~15,
  no.~5, p. 056013, 2018.

\bibitem{li2015eeg}
\BIBentryALTinterwordspacing
X.~Li, P.~Zhang, D.~Song, G.~Yu, Y.~Hou, and B.~Hu, ``Eeg based emotion
  identification using unsupervised deep feature learning,'' August 2015.
  [Online]. Available: \url{http://oro.open.ac.uk/44132/}
\BIBentrySTDinterwordspacing

\bibitem{thodoroff2016learning}
P.~Thodoroff, J.~Pineau, and A.~Lim, ``Learning robust features using deep
  learning for automatic seizure detection,'' in \emph{Machine learning for
  healthcare conference}.\hskip 1em plus 0.5em minus 0.4em\relax PMLR, 2016,
  pp. 178--190.

\bibitem{turk2019epilepsy}
{\"O}.~T{\"u}rk and M.~S. {\"O}zerdem, ``Epilepsy detection by using scalogram
  based convolutional neural network from eeg signals,'' \emph{Brain sciences},
  vol.~9, no.~5, p. 115, 2019.

\bibitem{RAGHU2020202}
\BIBentryALTinterwordspacing
S.~Raghu, N.~Sriraam, Y.~Temel, S.~V. Rao, and P.~L. Kubben, ``Eeg based
  multi-class seizure type classification using convolutional neural network
  and transfer learning,'' \emph{Neural Networks}, vol. 124, pp. 202--212,
  2020. [Online]. Available:
  \url{https://www.sciencedirect.com/science/article/pii/S0893608020300198}
\BIBentrySTDinterwordspacing

\bibitem{begleiter_1999}
\BIBentryALTinterwordspacing
H.~Begleiter, ``Eeg database,'' 1999. [Online]. Available:
  \url{https://kdd.ics.uci.edu/databases/eeg/eeg.data.html}
\BIBentrySTDinterwordspacing

\bibitem{spampinato2017deep}
C.~Spampinato, S.~Palazzo, I.~Kavasidis, D.~Giordano, N.~Souly, and M.~Shah,
  ``Deep learning human mind for automated visual classification,'' in
  \emph{Proceedings of the IEEE conference on computer vision and pattern
  recognition}, 2017, pp. 6809--6817.

\bibitem{li2020perils}
R.~Li, J.~S. Johansen, H.~Ahmed, T.~V. Ilyevsky, R.~B. Wilbur, H.~M. Bharadwaj,
  and J.~M. Siskind, ``The perils and pitfalls of block design for eeg
  classification experiments,'' \emph{IEEE Transactions on Pattern Analysis and
  Machine Intelligence}, vol.~43, no.~1, pp. 316--333, 2020.

\bibitem{palazzo2020correct}
S.~Palazzo, C.~Spampinato, J.~Schmidt, I.~Kavasidis, D.~Giordano, and M.~Shah,
  ``Correct block-design experiments mitigate temporal correlation bias in eeg
  classification,'' \emph{arXiv preprint arXiv:2012.03849}, 2020.

\bibitem{mukherjee2019cogni}
P.~Mukherjee, A.~Das, A.~K. Bhunia, and P.~P. Roy, ``Cogni-net: Cognitive
  feature learning through deep visual perception,'' in \emph{2019 IEEE
  International Conference on Image Processing (ICIP)}.\hskip 1em plus 0.5em
  minus 0.4em\relax IEEE, 2019, pp. 4539--4543.

\bibitem{fares2018region}
A.~Fares, S.~Zhong, and J.~Jiang, ``Region level bi-directional deep learning
  framework for eeg-based image classification,'' in \emph{2018 IEEE
  International Conference on Bioinformatics and Biomedicine (BIBM)}.\hskip 1em
  plus 0.5em minus 0.4em\relax IEEE, 2018, pp. 368--373.

\bibitem{zheng2020ensemble}
X.~Zheng, W.~Chen, Y.~You, Y.~Jiang, M.~Li, and T.~Zhang, ``Ensemble deep
  learning for automated visual classification using eeg signals,''
  \emph{Pattern Recognition}, vol. 102, p. 107147, 2020.

\bibitem{GRUTarnsformer}
Y.~Tao, T.~Sun, A.~Muhamed, S.~Genc, D.~Jackson, A.~Arsanjani, S.~Yaddanapudi,
  L.~Li, and P.~Kumar, ``Gated transformer for decoding human brain eeg
  signals,'' in \emph{2021 43rd Annual International Conference of the IEEE
  Engineering in Medicine \& Biology Society (EMBC)}.\hskip 1em plus 0.5em
  minus 0.4em\relax IEEE, 2021, pp. 125--130.

\bibitem{kavasidis2017brain2image}
I.~Kavasidis, S.~Palazzo, C.~Spampinato, D.~Giordano, and M.~Shah,
  ``Brain2image: Converting brain signals into images,'' in \emph{Proceedings
  of the 25th ACM international conference on Multimedia}, 2017, pp.
  1809--1817.

\bibitem{deng2009imagenet}
J.~Deng, W.~Dong, R.~Socher, L.-J. Li, K.~Li, and L.~Fei-Fei, ``Imagenet: A
  large-scale hierarchical image database,'' in \emph{2009 IEEE conference on
  computer vision and pattern recognition}.\hskip 1em plus 0.5em minus
  0.4em\relax Ieee, 2009, pp. 248--255.

\bibitem{VGG16}
K.~Simonyan and A.~Zisserman, ``Very deep convolutional networks for
  large-scale image recognition,'' \emph{arXiv preprint arXiv:1409.1556}, 2014.

\bibitem{ResNet}
\BIBentryALTinterwordspacing
K.~He, X.~Zhang, S.~Ren, and J.~Sun, ``Deep residual learning for image
  recognition,'' \emph{CoRR}, vol. abs/1512.03385, 2015. [Online]. Available:
  \url{http://arxiv.org/abs/1512.03385}
\BIBentrySTDinterwordspacing

\bibitem{MobileNets}
\BIBentryALTinterwordspacing
A.~G. Howard, M.~Zhu, B.~Chen, D.~Kalenichenko, W.~Wang, T.~Weyand,
  M.~Andreetto, and H.~Adam, ``Mobilenets: Efficient convolutional neural
  networks for mobile vision applications,'' \emph{CoRR}, vol. abs/1704.04861,
  2017. [Online]. Available: \url{http://arxiv.org/abs/1704.04861}
\BIBentrySTDinterwordspacing

\bibitem{EfficientNet}
\BIBentryALTinterwordspacing
M.~Tan and Q.~V. Le, ``Efficientnet: Rethinking model scaling for convolutional
  neural networks,'' \emph{CoRR}, vol. abs/1905.11946, 2019. [Online].
  Available: \url{http://arxiv.org/abs/1905.11946}
\BIBentrySTDinterwordspacing

\bibitem{haralick1973textural}
R.~M. Haralick, K.~Shanmugam, and I.~H. Dinstein, ``Textural features for image
  classification,'' \emph{IEEE Transactions on systems, man, and cybernetics},
  no.~6, pp. 610--621, 1973.

\bibitem{hu1962visual}
M.-K. Hu, ``Visual pattern recognition by moment invariants,'' \emph{IRE
  transactions on information theory}, vol.~8, no.~2, pp. 179--187, 1962.

\bibitem{ojala2002multiresolution}
T.~Ojala, M.~Pietikainen, and T.~Maenpaa, ``Multiresolution gray-scale and
  rotation invariant texture classification with local binary patterns,''
  \emph{IEEE Transactions on pattern analysis and machine intelligence},
  vol.~24, no.~7, pp. 971--987, 2002.

\bibitem{DBLP:journals/corr/abs-1904-09111}
\BIBentryALTinterwordspacing
E.~Vaineau, A.~Barachant, A.~Andreev, P.~C. Rodrigues, G.~Cattan, and
  M.~Congedo, ``Brain invaders adaptive versus non-adaptive {P300}
  brain-computer interface dataset,'' \emph{CoRR}, vol. abs/1904.09111, 2019.
  [Online]. Available: \url{http://arxiv.org/abs/1904.09111}
\BIBentrySTDinterwordspacing

\bibitem{ahmed2021object}
H.~Ahmed, R.~B. Wilbur, H.~M. Bharadwaj, and J.~M. Siskind, ``Object
  classification from randomized eeg trials,'' in \emph{Proceedings of the
  IEEE/CVF Conference on Computer Vision and Pattern Recognition}, 2021, pp.
  3845--3854.

\end{thebibliography}

\end{document}